\documentclass[letterpaper, 10 pt, journal]{ieeeconf}

\usepackage{amsmath} 
\usepackage{amssymb}  

\usepackage[dvipsnames]{xcolor}
\usepackage{soul}
\usepackage{graphicx}
\usepackage{url}

\let\labelindent\relax 

\usepackage[inline]{enumitem}
\usepackage{algorithm}
\usepackage{algpseudocode}

\usepackage{tabularray}
\usepackage{booktabs}

\usepackage{cite}

\title{\LARGE \bf
A Distributional Treatment of Real2Sim2Real for Object-Centric Agent Adaptation in Vision-Driven Deformable Linear Object Manipulation
}

\author{Georgios Kamaras$^{1,2}$ and Subramanian Ramamoorthy$^{1,3}$
\thanks{$^{1}$School of Informatics, The University of Edinburgh, EH8 9AB, UK. {\tt\small \{gkamaras, s.ramamoorthy\}@ed.ac.uk}%
}
\thanks{$^{2}$Corresponding author. Work  supported by the Engineering and Physical Sciences Research Council (EPSRC), as part of the CDT in RAS at Heriot-Watt University and The University of Edinburgh.%
}
\thanks{$^{3}$Work supported by a UKRI Turing AI World Leading Researcher Fellowship on AI for Person-Centred and Teachable Autonomy (grant EP/Z534833/1).%
}
\thanks{For the purpose of open access, the authors have applied a Creative Commons Attribution (CC BY) licence to any Author Accepted Manuscript version arising from this submission.}
}

\begin{document}

\maketitle
\thispagestyle{empty}
\pagestyle{empty}

\begin{abstract}
    We present an integrated (or end-to-end) framework for the Real2Sim2Real problem of manipulating deformable linear objects (DLOs) based on visual perception. Working with a parameterised set of DLOs, we use likelihood-free inference (LFI) to compute the posterior distributions for the physical parameters using which we can approximately simulate the behaviour of each specific DLO. We use these posteriors for domain randomisation while training, in simulation, object-specific visuomotor policies (i.e. assuming only visual and proprioceptive sensory) for a DLO reaching task, using model-free reinforcement learning. We demonstrate the utility of this approach by deploying sim-trained DLO manipulation policies in the real world in a zero-shot manner, i.e. without any further fine-tuning. In this context, we evaluate the capacity of a prominent LFI method to perform fine classification over the parametric set of DLOs, using only visual and proprioceptive data obtained in a dynamic manipulation trajectory. We then study the implications of the resulting domain distributions in sim-based policy learning and real-world performance.
\end{abstract}

\section{Introduction}
\label{sec:intro}

\begin{figure}[!t]
    \centering
    \includegraphics[width=1.0\columnwidth]{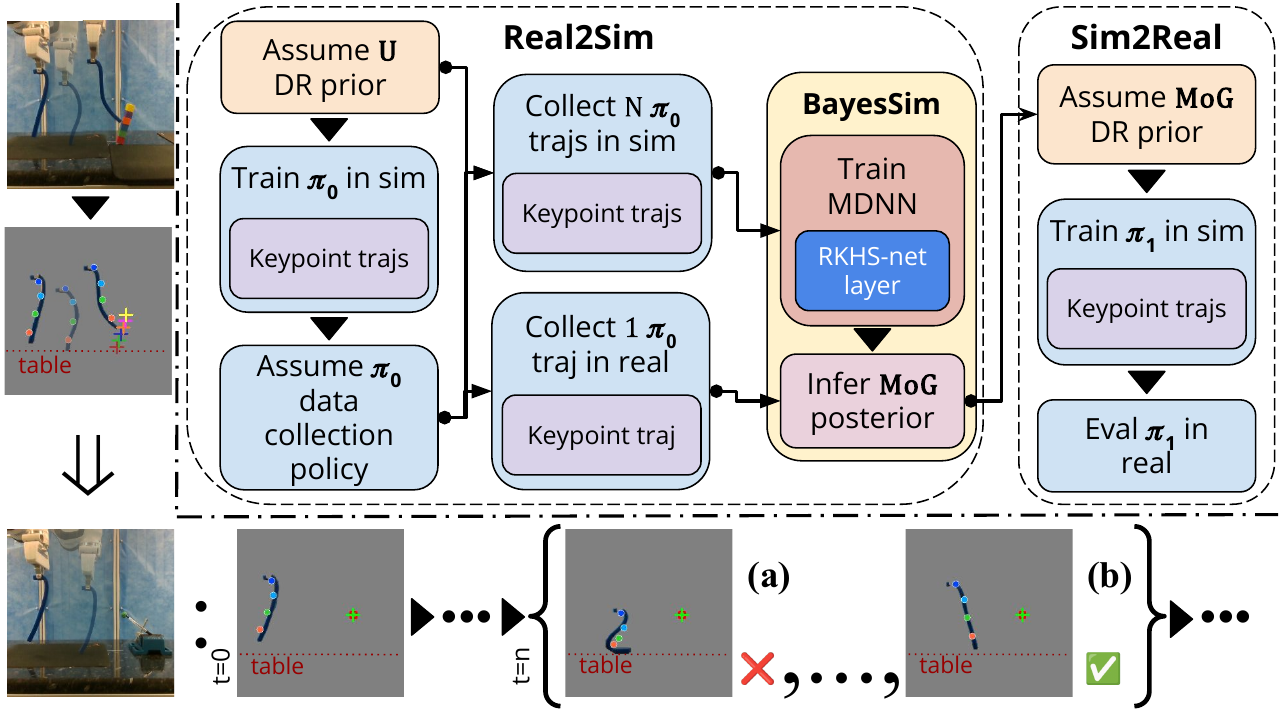}
    \caption{Overview of our Real2Sim2Real framework (top). We perform LFI for the posterior distribution $\hat{p}$ over system parameters (Real2Sim). We use $\hat{p}$ to perform domain randomisation while training a PPO agent to perform a DLO reaching task. We deploy and evaluate our sim-trained policy in the real world (Sim2Real). 
    We experiment with a DLO reaching task, which is an abstraction of more intricate physical interactions, such as whipping the top cube of a stack (left).
    We demonstrate strong object-centric agent adaptation relative to the underlying domain distribution $\hat{p}$. For example (bottom), an effective task policy should not drag the DLO on the table (a), but reach for the green target (cross) with a larger part of its body (b).}
    \label{fig:header-system-overview}
\end{figure}

Deformable linear object (DLO) manipulation is an active area of robotics research that deals with challenging tasks, such as lace tying, surgical suturing, and rope whipping~\cite{zhang2021robots, chi2024iterative, haiderbhai2024sim2real}.
Even when we know a DLO geometry, adapting a policy to its finer physical parameters is crucial to achieve precise physical interactions.
For this paper, let us consider a \emph{visuomotor} reaching task that seemingly requires less dexterity, but indicates readiness to scale up toward the aforementioned tasks. 
We hold a DLO above a table and want to guide its entire body toward a 2D visual target within a fixed time horizon, using visual and proprioceptive observations. Our objective can be quantified as minimising the overall distance of the DLO body to the target (Fig.~\ref{fig:header-system-overview}).

Let us think of how efficiently we would solve such a task; intuitively estimating DLO length and stiffness, from a few visual observations of manipulation trajectories, and adapting our actions to this estimation. 
An effective sequence of such actions would maximise whole-body convergence to the target for a given DLO parameterisation, while implicitly minimising adversarial factors, such as DLO drag on the table surface.
Achieving a similar level of robotic dexterity requires conditioning control policies to physical parameter estimations~\cite{kuroki2024gendom, zhang2024adaptigraph}, while dealing with the high dimensionality and nonlinearity of DLO states, challenges further exacerbated by the inherent noise in visual servoing~\cite{arriola2020modeling, yin2021modeling}.

Learning policies in simulation and then performing a \emph{Sim2Real} deployment~\cite{liang2024real, haiderbhai2024sim2real} follows the hypothesis that multiple simulated iterations of a task will robustify the learnt policy to parametric variations~\cite{peng2018sim}. This requires overcoming the \emph{``reality gap''} of robotic simulators, which is particularly problematic for soft objects. 
Thus, we first need a reliable \emph{Real2Sim} calibration~\cite{mehta2021user, liang2020learning} of the simulator parameters $\boldsymbol{\theta}$ to real-world parameters.
Then, for robust deployment, we account for any uncertainty about these parameters by exposing our learning algorithm to different $\boldsymbol{\theta}$ hypotheses.

Likelihood-free inference (LFI) deals with solving the inverse problem of probabilistically mapping real-world sensor observations to the respective simulation parameters $\boldsymbol{\theta}$ that are most likely to account for the observations.
This involves inferring the multimodal distribution $\hat{p}(\boldsymbol{\theta})$ from which we can sample sets of simulation parameters with a high probability of replicating the modelled real-world phenomenon~\cite{ramos2019bayessim}. Any uncertainty over $\boldsymbol{\theta}$ can be modelled as the variances associated with the modes.

This enables us to use $\hat{p}(\boldsymbol{\theta})$ as a sampler while performing Domain Randomisation (DR). DR aims to create a variety of simulated environments, each with randomised system parameters, and then to train a policy that works in all of them for a given control task objective. Assuming that the parameters of the real system is a sample in the distribution associated with the variations seen at training time, i.e. $\hat{p}(\boldsymbol{\theta})$, we can expect that a policy learnt in simulation under this regime will adapt to the real-world dynamics.

Recent literature has demonstrated how distributional embeddings, such as reproducing kernel Hilbert spaces~\cite{muandet2017kernel} of inferred low-dimensional keypoint trajectories, provide robustness to visual data noise and permutation invariance~\cite{antonova2022bayesian}, thus enabling robust Real2Sim calibration. On the Sim2Real side, we have seen fruitful model-based Reinforcement Learning (RL) approaches~\cite{zeng2021transporter, seita2021learning} to train deformable object manipulation policies in simulation and deploy them in the real world. However, we have yet to see an end-to-end \emph{Real2Sim2Real} system which combines the expressiveness of Bayesian inference with the flexibility of model-free RL~\cite{schulman2017proximal}.

In this paper, we make the following contributions.
\begin{enumerate}
    \item We propose an \textbf{end-to-end Real2Sim2Real framework} for robust \textbf{vision-based} DLO manipulation. 
    \item We examine the capacity of BayesSim~\cite{ramos2019bayessim} with distributional state embeddings to \textbf{finely classify} the physical properties of a DLO drawn from a parametric set.
    \item We study the implications of \textbf{different randomisation domains} for model-free RL policy learning in simulation and how this translates to \textbf{real-world performance}. 
\end{enumerate}
Our experiments show that for a parameterised set of (real) DLOs, an integrated distributional treatment of parameter inference, policy training, and zero-shot deployment enables inferring fine differences in physical properties and adapting RL agent behaviour to them.

\section{Background}
\label{sec:background}

\subsection{Likelihood-free Inference}
\label{subsec:lfi-prelim}

LFI treats a simulator as a black-box generative model $g$ with intractable likelihood, which uses its parameterisation $\boldsymbol{\theta}$ to generate an output $\mathbf{x}$ that represents the behaviour of the simulated system~\cite{papamakarios2016fast}. This process defines a likelihood function $p(\mathbf{x} \mid \boldsymbol{\theta})$, also called the forward model of the system, which we cannot evaluate, but we can sample from by running the simulator.
To overcome the ``reality gap", we want to solve the inverse problem, which is mapping real observations $\mathbf{x}^r$ to the parameters $\boldsymbol{\theta}$ that are most likely to replicate them in simulation; let $\mathbf{x}^s = g(\boldsymbol{\theta})$. This defines the problem of approximating the posterior $\hat{p}(\boldsymbol{\theta} \mid \mathbf{x}^s, \mathbf{x}^r)$.

BayesSim~\cite{ramos2019bayessim} approximates the posterior $\hat{p}(\boldsymbol{\theta} \mid \mathbf{x} = \mathbf{x}^r)$ by learning the conditional density function $q_{\phi}(\boldsymbol{\theta} \mid \mathbf{x})$, parameterised by $\boldsymbol{\phi}$. Conditional density is modelled as a mixture of Gaussians (MoG), which can be approximated by mixture density neural networks (MDNN)~\cite{bishop1994mixture}. The inputs $\mathbf{x}$ can be high-dimensional state-action trajectories, summary statistics $\psi(\cdot)$ of trajectory data, or kernel mappings.

BayesSim first requires training $q_{\phi}(\boldsymbol{\theta} \mid \mathbf{x})$ on a dataset of $N$ pairs $(\boldsymbol{\theta}_n, \mathbf{x}_n)$, where the parameters $\boldsymbol{\theta}_n$ are drawn from a \emph{proposal prior} $\Tilde{p}(\boldsymbol{\theta})$ and the observation trajectories $\mathbf{x}_n$ are generated by running $g(\boldsymbol{\theta}_n)$ for the duration of a simulated episode and collecting the respective state-action pairs.

Then, given a single real world trajectory $\mathbf{x}^r$, BayesSim estimates the posterior as:
\begin{equation}
    \hat{p}(\boldsymbol{\theta} \mid \mathbf{x} = \mathbf{x}^r) \propto \frac{p(\boldsymbol{\theta})}{\Tilde{p}(\boldsymbol{\theta})} q_{\phi}(\boldsymbol{\theta} \mid \mathbf{x} = \mathbf{x}^r) \text{,}
\end{equation}
which allows the flexibility (`likelihood-free') for a desirable prior $p(\boldsymbol{\theta})$ which is different than the proposal prior. If $\Tilde{p}(\boldsymbol{\theta}) = p(\boldsymbol{\theta})$, then $ \hat{p}(\boldsymbol{\theta} \mid \mathbf{x} = \mathbf{x}^r) \propto q_{\phi}(\boldsymbol{\theta} \mid \mathbf{x} = \mathbf{x}^r)$.

\subsection{Domain Randomisation}
\label{subsec:dr-prelim}

Domain randomisation using wide uniform priors~\cite{tobin2017domain}, combined with the inherent instability of training RL algorithms, has been shown to not provide the robustness expected in certain tasks~\cite{peng2018sim, possas2020online}. 
Using LFI, we obtain a posterior $\hat{p}(\boldsymbol{\theta} \mid \mathbf{x} = \mathbf{x}^r)$ which we expect to be qualitatively more precise, i.e. narrower than the uniform prior, and more accurate, i.e. dense around the system's true parameters.

The goal of RL is to maximise the expected sum of future discounted rewards (by $\gamma$) following a policy $\pi_{\boldsymbol{\beta}}(\mathbf{a}_t \mid \mathbf{s}_t)$, parameterised by $\boldsymbol{\beta}$ and sampling action $\mathbf{a}_t$ given state $\mathbf{s}_t$.
We formulate DR over the RL problem as maximising the joint objective:
\begin{equation}
    \mathcal{J}(\boldsymbol{\beta}) = \mathbb{E}_{\boldsymbol{\theta}} \Biggl[ \mathbb{E}_{\boldsymbol{\eta}} \Biggl[ \sum_{t=0}^{T-1} \gamma^{(t)} r(\mathbf{s}_t, \mathbf{a}_t) \mid \boldsymbol{\beta} \Biggr] \Biggr] \text{,}
\end{equation}
with respect to policy's $\boldsymbol{\beta}$, where $\boldsymbol{\theta} \sim \hat{p}(\boldsymbol{\theta} \mid \mathbf{x} = \mathbf{x}^r)$ and $\boldsymbol{\eta}=\{\mathbf{s}_t, \mathbf{a}_t, \mathbf{o}_t\}_{t=0}^{T-1}$ the history of observation $\mathbf{o}_t$, action $\mathbf{a}_t$ pairs over time horizon $T$. In this paper, we consider $\mathbf{s}_t \equiv \mathbf{o}_t$.

\subsection{Kernel Mean Embeddings \& Neural Approximation}
\label{subsec:rkhs-bg}

Kernel mean embeddings map distributions into infinite-dimensional feature spaces, thus representing probability distributions without information loss to a space that enables useful mathematical operations.
This happens by turning the expectation operator into a reproducing kernel Hilbert space (RKHS) inner product, which has linear complexity in the number of training points~\cite{muandet2017kernel}.

Let $X$ be a random variable in sample space $\mathcal{X}$, $\mathcal{F}$ be an RKHS with kernel $k(\mathbf{x}, \mathbf{x}')$, and data points $\mathbf{x} \in \mathcal{X}$. $\mathcal{F}$ is a Hilbert space of functions $f : \mathcal{X} \rightarrow \mathbb{R}$ with the inner product $\langle \cdot , \cdot \rangle_{\mathcal{F}}$~\cite{ghojogh2021reproducing}.
Given a distribution of functions $P(X)$ and a mean embedding map $\mu_X \in \mathcal{F}$, we can recover expectations of all functions, as $\mathbb{E}_X [f(x)] = \langle \mu_X, f \rangle_{\mathcal{F}}, \forall f \in \mathcal{F}$. 
By Riesz representation theorem~\cite{akhiezer2013theory}, $\mu_X$ exists and is unique.
A common choice for the $\mu_X$ mapping is the Radial Basis Function (RBF) kernel, with parameterisable length scale $\sigma$.

Following~\cite{antonova2022bayesian}, we can define the empirical kernel embedding using i.i.d. samples $\mathbf{x}_1,...,\mathbf{x}_N$ from $P(X)$, as:
\begin{equation}
\label{eq:kernel-mean-embed}
    \hat{\mu}_X := \frac{1}{N} \sum_{n=1}^{N} \phi(\mathbf{x}_n) \text{.}
\end{equation}
With \emph{kernel trick} we avoid operating on infinite-dimensional implicit maps $\phi(\mathbf{x})$, and instead operate on a finite-dimensional Gram matrix $K_{ij}=k(\mathbf{x}_i, \mathbf{x}_j),i,j=1...N$~\cite{ghojogh2021reproducing}, which, however, is computationally expensive for large datasets ($\mathcal{O}(n^2)$).

Following~\cite{ramos2019bayessim} (Eq.~12-15) and~\cite{antonova2022bayesian} (Eq.~6) derivations, we can use \emph{Random Fourier Features} (RFF)~\cite{rahimi2007random} to approximate a shift-invariant kernel $k(\mathbf{x}, \mathbf{x}')$ by a dot product $\hat{\phi}(\mathbf{x})^T\hat{\phi}(\mathbf{x}')$, which is a more scalable approximation of $K$. 
$\hat{\phi}(\mathbf{x})$ is a finite-dimensional approximation of $\phi(\mathbf{x})$, with randomised features of the form $\cos(\boldsymbol{\omega}_{m}^T\mathbf{x}+\mathbf{b}_m)$ (cos-only) or $\langle\cos(\sigma^{-1}\circ\boldsymbol{\omega}_{m}^T\mathbf{x}),\sin(\sigma^{-1}\circ\boldsymbol{\omega}_{m}^T\mathbf{x})\rangle$ (cos-sin)~\cite{ramos2023method}. 

Following~\cite{antonova2022bayesian}, we use the RFF feature approximation to construct the kernel mean embedding of vectors \eqref{eq:kernel-mean-embed} that can benefit from a distributional representation.
This embedding approach can be integrated into a learning architecture in a fully differentiable manner, by constructing a neural network layer (RKHS-Net) that obtains random samples for frequencies $\boldsymbol{\omega}$, biases $\mathbf{b}$, and optionally $\sigma$ (e.g., in cos-sin), and adjusts them during training by computing gradients w.r.t. the overall architecture loss.

\section{Method}
\label{sec:method}

\begin{algorithm}[t]
\caption{Real2Sim2Real for DLO manipulation}
\label{alg:real2sim2real-bsim}
\begin{algorithmic}[1]
    \State \textbf{Given:} $N_{\text{LFI}}$: inference iterations
    \State Assume uniform proposal prior $\Tilde{p}(\boldsymbol{\theta}) \approx \mathit{U}$
    \State Assign reference prior $p_0 \gets \Tilde{p}$
    \State Train initial policy $\pi_{\boldsymbol{\beta}_0}(\mathbf{a}_t \mid \mathbf{s}_t)$, $\boldsymbol{\theta} \sim p_0$
    \State Run $1$ $\pi_{\boldsymbol{\beta}_0}$ rollout in the real env to collect $\mathbf{x}^r$
    \State \textbf{// 1. Real2Sim DLO parameter inference (LFI)}
    \State $i \gets 0$
    \While{$i < N_{\text{LFI}}$} \label{alg:r2s2r:line:lfi-iter}
        \State $\{(\boldsymbol{\theta}, \mathbf{x}^s)\}^N \gets$ Run $N$ $\pi_{\boldsymbol{\beta}_0}$ rollouts in sim, $\boldsymbol{\theta} \sim p_i$
        \State Train $q_{\phi}$ over $\{(\boldsymbol{\theta}, \mathbf{x}^s)\}^N$
        \State $\hat{p}(\boldsymbol{\theta} \mid \mathbf{x} = \mathbf{x}^r) \propto p_i(\boldsymbol{\theta}) \mathbin{/} \Tilde{p}(\boldsymbol{\theta}) q_{\phi}(\boldsymbol{\theta} \mid \mathbf{x} = \mathbf{x}^r)$
        \State $i \gets i + 1$
        \State Update reference prior $p_i \gets \hat{p}$ \label{alg:r2s2r:line:p-update}
    \EndWhile
    \State \textbf{// 2. Policy training in sim}
    \State Train policy $\pi_{\boldsymbol{\beta}_1}(\mathbf{a}_t \mid \mathbf{s}_t)$, $\boldsymbol{\theta} \sim \hat{p}$
    \State \textbf{// 3. Sim2Real policy deployment}
    \State Evaluate $\pi_{\boldsymbol{\beta}_1}$ in the real env by running $1$ $\pi_{\boldsymbol{\beta}_1}$ rollout
\end{algorithmic}
\end{algorithm}

\subsection{Keypoint detection from segmentation images}
\label{subsec:real2sim2real-seg-img-n-kps}

Our perception module receives an RGB image as input, in which it detects the environment objects of interest, i.e. the controlled DLO and the task's target. We use detection masks to create segmentation images, through which we track our task's features for parameter inference and policy learning.

We use segmentation images to efficiently learn an unsupervised keypoint detection model, using the \emph{transporter} method~\cite{kulkarni2019unsupervised}, as implemented by~\cite{li2020causal}. The resulting keypoints, although temporally consistent, are inferred in every timeframe, creating problems such as pixel position noise and permutations~\cite{antonova2022bayesian}, particularly when working with real-world image data of deformable objects.
The lack of permutation invariance hinders the creation of reward functions specific to DLO parts, such as the tip~\cite{lim2022real2sim2real, chi2024iterative}. Focussing on whole-body guidance abstracts this challenge and raises the importance of inferred posteriors to implicitly learn successful behaviours using less elaborate reward functions.

\subsection{Real2Sim with likelihood-free inference}
\label{subsec:real2sim-bayessim}

Considering our prefaced DLO visuomotor reaching task, we want to learn a policy in simulation and deploy it in the real world, without any further fine-tuning. The Real2Sim part of our work deals with calibrating physical parameters that induce the deformable object's behaviour, such as its dimensions and material behaviour, which are difficult to tune manually. Details such as camera and workspace object placement, controller stiffness parameters, etc. are beyond the scope of our work, although they can be incorporated as additional tunable parameters in extensions of our method.

For our inference and domain randomisation experiments, we define a physical parameter vector $\boldsymbol{\theta} = \langle l, E \rangle$, where $l$ denotes the length of our DLO and $E$ denotes its Young's modulus. Thus, we want to infer a joint posterior $\hat{p}$, which contains information on both the dimensions and the material properties of our deformable object. For this, we use the BayesSim-RKHS variant~\cite{antonova2022bayesian}.

Following Algorithm~\ref{alg:real2sim2real-bsim}, we begin by assuming a uniform proposal prior $\Tilde{p}(\boldsymbol{\theta})$, which we use to initialise our reference prior $p_0 = \Tilde{p}$. We then perform domain randomisation as $\boldsymbol{\theta} \sim p_0$, while training in simulation an initial policy $\pi_{\boldsymbol{\beta}_0}$ for our task. We execute a rollout of $\pi_{\boldsymbol{\beta}_0}$ in the real world environment, to collect a real trajectory $\mathbf{x}^r$ while manipulating a specific DLO. We perform LFI through \emph{multiple} BayesSim iterations~\cite{possas2020online, antonova2022bayesian}. In each inference iteration $i$, we use $\pi_{\boldsymbol{\beta}_0}$ as a data collection policy, running $N$ rollouts of $\pi_{\boldsymbol{\beta}_0}$ in simulation to collect a dataset $\{(\boldsymbol{\theta}, \mathbf{x}^s)\}^N, \boldsymbol{\theta} \sim \Tilde{p}$, on which we train our BayesSim conditional density function $q_{\phi}$. We use our $q_{\phi}$ and $\mathbf{x}^r$ to compute the posterior $\hat{p}(\boldsymbol{\theta} \mid \mathbf{x} = \mathbf{x}^r)$. We then update our reference prior $p_i = \hat{p}$ and loop again.

We can now adapt our domain randomisation distribution to the latest inferred posterior $\hat{p}$, and we proceed to retrain our task policy $\pi_{\boldsymbol{\beta}_1}$. Our hypothesis is that by sampling $\boldsymbol{\theta}$ from our object-specific posterior, we will get faster $\pi_{\boldsymbol{\beta}_1}$ convergence to consistently successful behaviour, and that running a rollout of $\pi_{\boldsymbol{\beta}_1}$ in the real-world environment we will collect higher cumulative reward for the specific DLO within the task episode's horizon.

\begin{figure}[t]
    \centering
    \includegraphics[width=1.0\columnwidth]{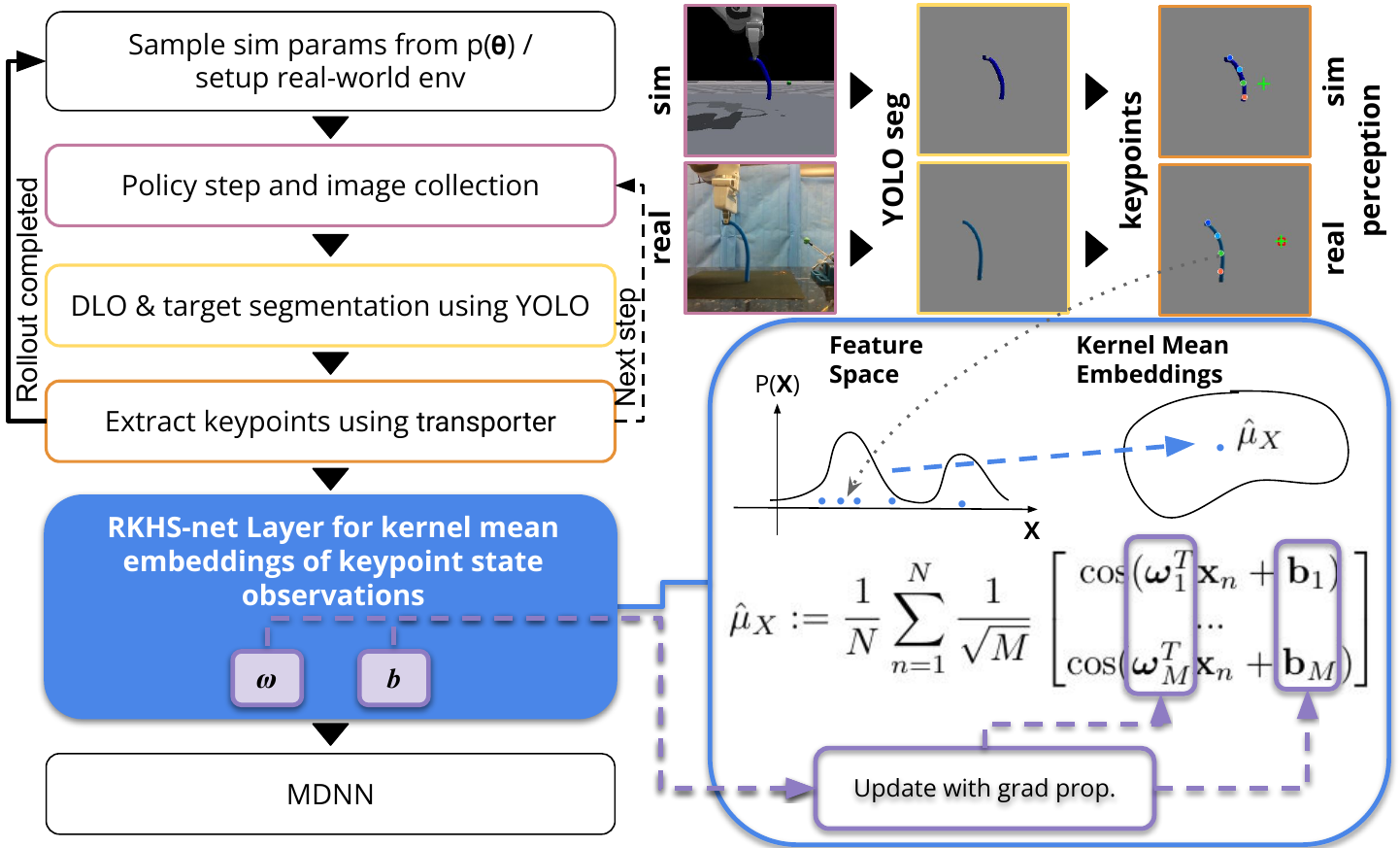}
    \caption{Overview of our policy rollout and trajectory perception method. In each policy step we extract keypoints from segmentation images. Once all rollouts are completed, for each keypoint trajectory we compute the cos-only kernel mean embeddings of each step's keypoints using the RKHS-net layer (BayesSim-RKHS). Margins colour coding indicates association between sample images and algorithm parts. Illustration inspired by~\cite{antonova2022bayesian}.}
    \label{fig:perc-to-rkhs}
\end{figure}

\subsection{Keypoint trajectories in RKHS}

We use an RKHS-net layer~\cite{antonova2022bayesian} to construct a distributional state representation (Fig.~\ref{fig:perc-to-rkhs}), which is provably robust to visual noise and uncertainty due to inferred keypoint permutations. Intuitively, through our kernel mean embeddings~\cite{muandet2017kernel} we pull data from the input space (keypoint trajectory) to the feature space (RKHS). The explicit locations of the pulled points are not necessarily known or might be uncertain, but the relative similarity (inner product) of the pulled data points is known in the feature space~\cite{ghojogh2021reproducing}. This gives us noise robustness and permutation invariance, and it generally means that functional representations that are similar to each other are embedded closer within the RKHS, and thus they are more likely to be classified as similar.

The input of the RKHS-net layer is a vector of samples of a distribution. In this work, we compute the distributional embedding of a noisy keypoint trajectory $\mathbf{x}=(\mathbf{x_1},...,\mathbf{x_n})$, each keypoint defined in a $2$D RGB pixel space. 

\subsection{Policy Learning and Sim2Real Deployment}
\label{subsec:real2sim-param-inference}

We use Proximal Policy Optimisation (PPO)~\cite{schulman2017proximal} as our reference model-free \emph{on-policy} RL algorithm. PPO maximises the expected reward with a clipped surrogate objective to prevent large updates. It samples actions from a Gaussian distribution, which is parameterised during policy learning.
We learn policies in simulation, performing domain randomisation using our LFI \emph{reference prior} $p$ as a task parameterisation sampler, where $p$ is either the default uniform distribution $\mathit{U}$, or an inferred MoG posterior (Alg.~\ref{alg:real2sim2real-bsim}, l.~\ref{alg:r2s2r:line:p-update}).

\section{Experimental Setup}
\label{sec:experiments}

Our experiments address the following questions.
\begin{enumerate}
    \item Can our inferred posteriors \textbf{describe different physical parameterisations of similarly shaped DLOs}?
    \item What is \textbf{the impact of these posteriors on domain randomisation} using the respective MoG?
    \item How does this translate into \textbf{object-centric agent performance adaptation} during real-world deployment?
\end{enumerate}

\subsection{Reference Task}
\label{subsec:reference-task}

To evaluate the parameter inference performance of our setup, and the implications of the inferred parameter posterior $\hat{p}(\boldsymbol{\theta})$ for policy learning and deployment, we design a visuomotor reaching task that studies the guidance of the entire DLO body towards a visual target.

We set up our task with a robot arm that picks up the DLO from a designated position on the table $\mathbf{x}_0$ by grasping it near one of its tips, and raises it to a designated height $h_0$. Both $\mathbf{x}_0$ and $h_0$, as well as our initial DLO pose, remain fixed for all of our simulation and real experiment iterations. The execution of our policy begins immediately after the object has been picked up and raised to $h_0$. Thus, the dynamic nature of our task is constituted by the underactuated manner in which we control our DLO, which dangles from one of its tips, as well as the momentum forces (drag, inertia) acting on its body ever since it was raised from the table.

The positions of the DLO and the target are tracked in the 2D pixel space of an RGB image. The DLO is tracked by an inferred cluster of $4$ keypoints, extracted over a segmentation image, and the target with the computed pixel space median of its own segmentation mask, which we treat as an artificial $5$th keypoint.
In both our parameter inference (real2sim) and policy training and deployment (sim2real) experiments, we position-control our Panda end-effector (EEF) by commanding its Cartesian pose. Our simulator uses the IsaacGym attractors implementation, whereas in the real world we use a Cartesian impedance controller~\cite{luo2024serl}.

\subsection{Simulation \& Hardware setup}
\label{subsec:sim-hw-setup}

We setup our simulation in IsaacGym~\cite{makoviychuk2021isaac}. Our simulated environment contains a Franka Emika Panda 7-DoF robot arm equipped with a parallel gripper, which is on top of a tabletop workspace. On top of this workspace, a blue DLO exists, implemented as a tetrahedral grid,
and simulated using the corotational finite element method of the FleX physics engine. A small green spherical object is our task's reaching target. Across all our parameter inference and policy learning experiments, the simulated DLO is parameterised in the $[1e3, 5e4]$ (Pa) range for Young's modulus and the $[195, 305]$ (mm) range for length. Thus, \emph{median} of our simulated parameter space is $\mu = (2.55e4 \; \text{Pa}, \; 250\text{mm})$.

Our real-world setup closely replicates the simulation setup. We extend an open-source sample-efficient robotic RL infrastructure~\cite{luo2024serl} for the purpose of evaluating PPO policies~\cite{stable-baselines3}. For visual observations, we use a RealSense D435i camera capturing $60$ fps, mounted to view our workspace from the right side. To avoid the overhead of a more elaborate sim and real camera calibration, we select our real camera placement so that the captured images qualitatively approximate the respective sim images similar to~\cite{matl2020inferring, antonova2022bayesian}.

For our experiments, we manufacture $4$ blue DLOs using Shore hardness A-40 (\underline{DLO-0} with len. $\text{200mm}$), $\text{00-20}$ (\underline{DLO-1} with len. $\text{200mm}$ and \underline{DLO-3} with len. $\text{290mm}$) and $\text{00-50}$ (\underline{DLO-2} with len. $\text{270mm}$) silicone polymers~\cite{liao2020ecoflex}. 
Similarly to simulation, our real DLOs are shaped as grids, with negligibly small height and width at $15\text{mm}$ each. 
Table~\ref{tab:dlos} summarises our DLO indexes, parameterisations, relative hardness descriptions, and mass. Items are sorted primarily on increasing length and then on increasing softness (as \emph{medium soft $\rightarrow$ soft $\rightarrow$ extra soft}).
We will be referencing DLOs using their index, or as ``$\langle \text{Length} \rangle ; \langle \text{Shore Hardness} \rangle$''.

The real green target ball is made of plasticine. It is simulated with a fixed rigid sphere of radius $r=1$cm. Since DLO-target physical interactions are irrelevant to our task, this reality gap of the target material is also irrelevant.

\begin{table}[t]
\begin{center}
\caption{Real DLO indexes and parameterisations.}
\label{tab:dlos}
\begin{tabular}{c c c c c} 
    \textbf{DLO idx.} & 0 & 1 & 2 & 3 \\
    \midrule
    \textbf{Length} & 200mm & 200mm & 270mm & 290mm  \\
    \midrule
    \textbf{Shore} & A-40 & 00-20 & 00-50 & 00-20 \\
    \textbf{Hardness} & \emph{(medium soft)} & \emph{(extra soft)} & \emph{(soft)} & \emph{(extra soft)} \\ 
    \midrule
    \textbf{Mass} & $47.47$g & $42.94$g & $57.8$g & $63.94$g 
\end{tabular}
\end{center}
\end{table}

\subsection{Perception setup}
\label{subsec:perc-setup}

We collect visual observations using the real-time RGB image stream of our side-view camera. For computational efficiency, we focus our observations on our blue DLO and the green target ball using segmentation images. For this, we fine-tune the segmentation task version of YOLOv8.2~\cite{redmon2016you}. 
We use a dataset of $183$ manually labelled workspace images, pre-processed through auto-orientation and resizing to $256\times256$, and augmented using random Gaussian blur and salt-and-pepper noise, totalling $439$ images. With this dataset, we train the open source YOLO weights for another $64$ epochs. 

We reduce the dimensionality of our visual observations by applying a learnt model of $4$ keypoints to each segmentation image in real time. We train this keypoint model using a dataset of $1500$ random policy rollouts of our simulated control task, while sampling our system parameters $\boldsymbol{\theta}$ from a uniform prior. In this data set, we also incorporate a small number of real-world policy rollouts, and we train for $50$ epochs. These keypoints are our DLO observation vector.

\subsection{Domain Randomisation setup}
\label{subsec:dr-setup}

In practice, we see that in popular robotics simulators, such as IsaacGym, re-parameterising an existing deformable object simulation to change physical properties such as stiffness would require to re-initialise the whole simulation. This makes it difficult to integrate such simulators in an RL environment that follows the well-established \emph{gym} style~\cite{towers2024gymnasium}.

Thus, to perform DR for our RL task, we train our policies in \emph{vectorised} environments~\cite{stable-baselines3} and launch parallel instances of our simulation, each with a different $\boldsymbol{\theta}$, sampled by our current reference prior $p(\boldsymbol{\theta})$. To further robustify the Sim2Real transfer of our policies and minimise the need for camera calibration, we introduce a small randomness in our camera and target object positions by uniformly sampling $(x,y,z)$ offsets in the $(\pm0.025,\;\pm0.025,\;\pm0.025)$ (m) and $(\pm0.02,\;0,\;\pm0.02)$ (m) ranges, respectively.

We empirically chose to launch $12$ concurrent environments, from $12$ respective domain samples, to manage our computationally intensive deformable object simulations. This places the descriptiveness of our inferred posteriors at the forefront of our experimental study, since inaccuracy and imprecision of the inferred MoGs can have worse consequences in small-data experiments. We sample our MoGs in a \emph{low-variance} method, thus each component is likely to contribute to our set of domain samples.

\subsection{Control setup}
\label{subsec:ctrl-setup}

We command our EEF by sending it cartesian pose commands ($7$D vectors). For our control experiments, we constrain our EEF motion into two dimensions, moving only along the $x$ and $z$ axes by controlling the respective deltas. Thus, our sampled policy actions are $2$D $\langle dx, dz \rangle$ vectors that we sample in the $[-0.06, 0.06]$ (m) range to maintain smooth EEF transitions. Our $12$D observations are constructed by flattening and concatenating our EEF's $\langle x, z \rangle$ $2$D position vector (proprioception) and the $5 \times 2$D visual observation, which are the positions of our DLO and target keypoints.
 
In each state $s_t$, we compute the distance $d_t$ of our DLO keypoint cluster to the target using the Frobenius norm of the respective distance matrix. Our sparse reward function uses this $d_t$ and a pixel space distance threshold $d_{\text{thresh}} = 1.5$ to count the reward $r_t$. Thus, whenever $d_t \le d_{\text{thresh}}$, our respective reward is the distance scaled to $[0.0, 1.0]$, denoted as $r_t = 1.0 - (d_t / d_{\text{thresh}})$. The episode ends if $r_t \geq 0.75$.

For safety reasons, we restrict our EEF motion within the $\langle x, y, z \rangle \in \langle [0.275, 0.6], [-0.1, 0.1], [0.1, 0.5] \rangle$ (m) world frame coordinates, with the robot arm based on $(0, 0, 0)$. Whenever the EEF leaves this designated workspace, the episode ends as a failure with a reward of $-1$.

\subsection{Parameter Inference \& Policy Algorithms training}
\label{subsec:training-setup}

For parameter inference, we implement the BayesSim-RKHS variant~\cite{antonova2022bayesian} (Fig.~\ref{fig:perc-to-rkhs}). Our RKHS-net layer embeds the $5$ observed keypoints using $500$ random Fourier features and cos-only components. Our MDNN models $4$ mixture components with full covariance and uses $3$ fully connected layers of $1024$ units each. We use Adam optimiser with a learning rate of $1e$$-6$. As in Alg.~\ref{alg:real2sim2real-bsim}, line~\ref{alg:r2s2r:line:lfi-iter}, we approximate our posterior $\hat{p}(\boldsymbol{\theta})$ through $15$ iterations, each increasing the training set with $100$ more trajectories, whose parameters $\boldsymbol{\theta}$ have been sampled using the latest posterior.

For policy learning, we use the Stable Baselines3~\cite{stable-baselines3} PPO implementation. To maintain the generality of our proposed methodology, we keep the default implementation hyperparameters, which reflect the hyperparameters originally proposed in the seminal work~\cite{schulman2017proximal}. We train for $120,000$ total steps, with a maximum duration of episodes of $16$ steps, and a batch size of also $16$. We empirically tune our real impedance controller damping and stiffness so that $16$ real-world steps approximate the respective $16$ simulation steps.

\section{Results}
\label{sec:results}

\begin{figure}[t]
    \centering
    \includegraphics[width=1.0\columnwidth]{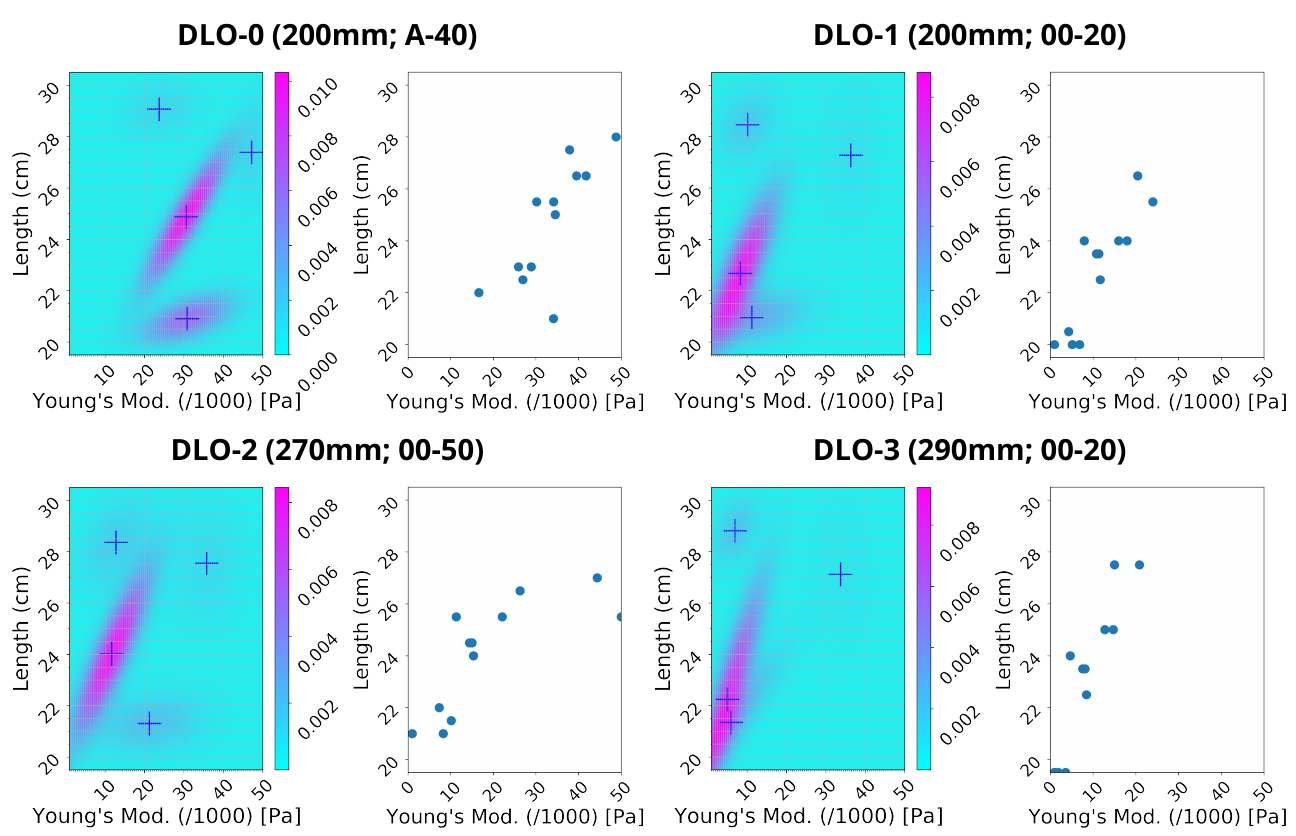}
    \caption{Inferred MoG posterior heatmaps and the induced domain samples when each MoG is used for DR. 
    MoG component means are displayed in blue crosses and colorbar quantifies likelihood.}
    \label{fig:mog-posterior-for-dr}
\end{figure}

\subsection{Different Physical Parameterisations in LFI \& DR}
\label{subsec:real2sim-res--lfi}

Figure~\ref{fig:mog-posterior-for-dr} illustrates our Real2Sim performance using 2D MoG posterior heatmaps, along with the respective scatter plots of the $12$ resulting domain samples. Each MoG has $4$ components, whose mean, variance, and mixture coefficient are parameterised during inference. The tightness and spread of the posteriors serve as a qualitative indication of the precision of the inference. The crosses that represent the means of different Gaussians capture alternative hypotheses of the observed DLO's parameterisation. 

We see that for our set of DLOs, BayesSim-RKHS \textbf{correctly classifies the varied softness} (Young's modulus); however, it \textbf{struggles to cleanly classify the varied dimension} (length). This is indicated by the MoG variance across each parameter's axis and the respective spread of the component means, which inherently encode any uncertainty about the parameter estimation.

\subsection{Uncertainty over Parameter Estimation \& DR}
\label{subsec:real2sim-res--dr}

Following the above property of MoG posteriors, we observe how \textbf{the relative certainty (\emph{sharpness}) of a posterior} along the dimension of a variable \textbf{results in proportionately tightly clustered domain samples} along this dimension. Thus we see domain samples being mainly spread along the less certain length dim. of posteriors. This is particularly visible for the notably similar posteriors of the very soft DLOs 1 and 3. 
For DLOs 0 and 2, we see that some of the \emph{alternative hypotheses} (component means) for $\boldsymbol{\theta}$ contribute to our low-variance sampling, due to the relative magnitude of the respective mixture coefficients. This contribution is evident in the spread of the respective domain samples.

\subsection{Object-centric Agent Performance}
\label{subsec:policy-learn-sim2real-res}

\begin{figure}[t]
    \centering
    \includegraphics[width=0.9\columnwidth]{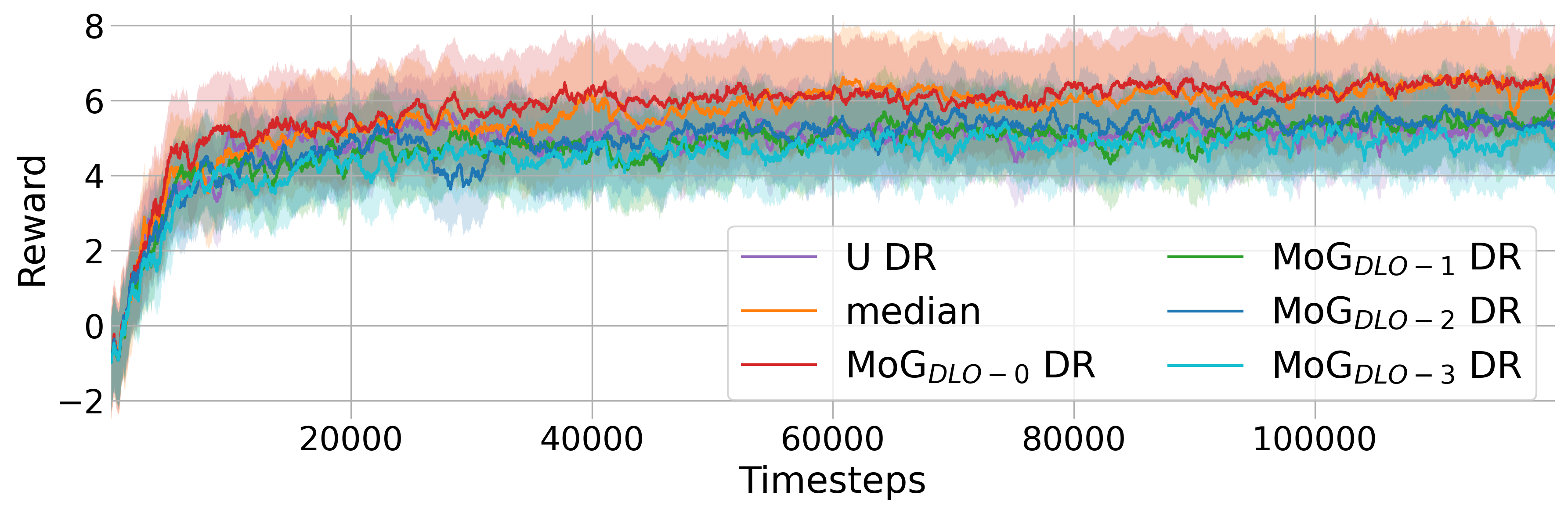}
    \caption{Learning curves of PPO agent training when performing DR using different domain distributions.}
    \label{fig:rew-per-train-iter}
\end{figure}

We train (Fig.~\ref{fig:rew-per-train-iter}) $4$ PPO policies using our $4$ inferred MoG posteriors (Fig.~\ref{fig:mog-posterior-for-dr}) for domain randomisation. We also train a policy by performing DR over a uniform distribution (\setulcolor{purple}\ul{$\text{PPO-}\mathit{U}$}) and a policy which assumes that the simulated DLO is parameterised according to the median of our parameter space (\setulcolor{orange}\ul{$\text{PPO-}\mu$}). We evaluate these $6$ policies in each one of our $4$ real-world DLOs, repeating each evaluation $4$ times for a total of $96$ sim2real policy deployments. 
Similarly, we also evaluate in simulation over a DLO-$\mu$ parameterised per our parameter space median $\mu$.

\begin{figure*}[t]
    \vspace{5pt}
    \centering
    \includegraphics[width=0.955\textwidth]{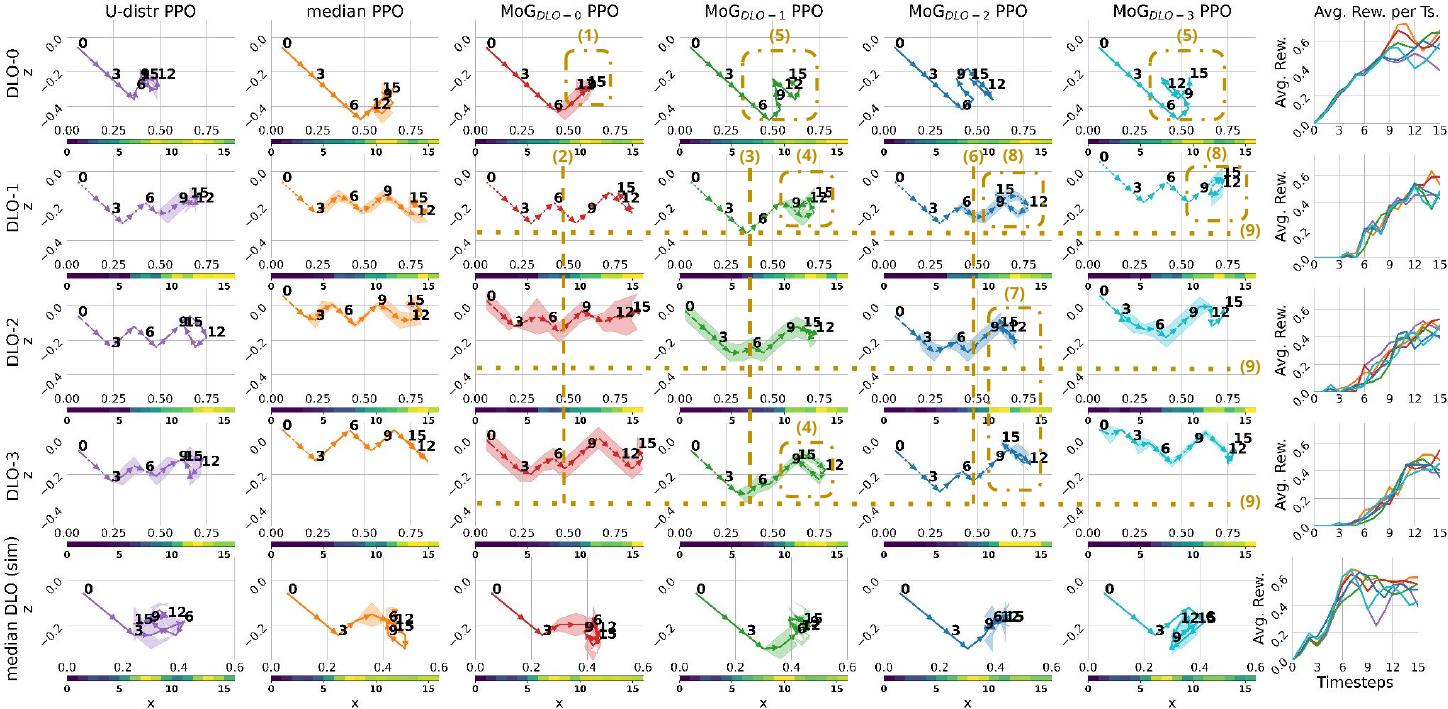}
    \caption{EEF trajectories during the deployment of $6$ policies trained using different domain distributions (col. 1-6) in real world (rows 1-4; $4$ real DLOs) and in simulation (row 5; median DLO).
    We repeat each deployment $4$ times and average the measured accumulation of commanded EEF translations along the $x$ and $z$ axes (commanded \emph{actions}, $\langle dx, dz \rangle$), with shaded regions reporting standard deviation. 
    Colorbars illustrate DLO keypoints pixel-distance to the target $d_t$, reversing reward function for $d_{\text{thresh}} = 1.5$. The rightmost column shows the averaged rewards per timestep for the respective row's trajectories.}
    \label{fig:eef-deltas-per-eval-steps}
\end{figure*}

\begin{figure*}[t]
    \centering
    \includegraphics[width=0.925\textwidth]{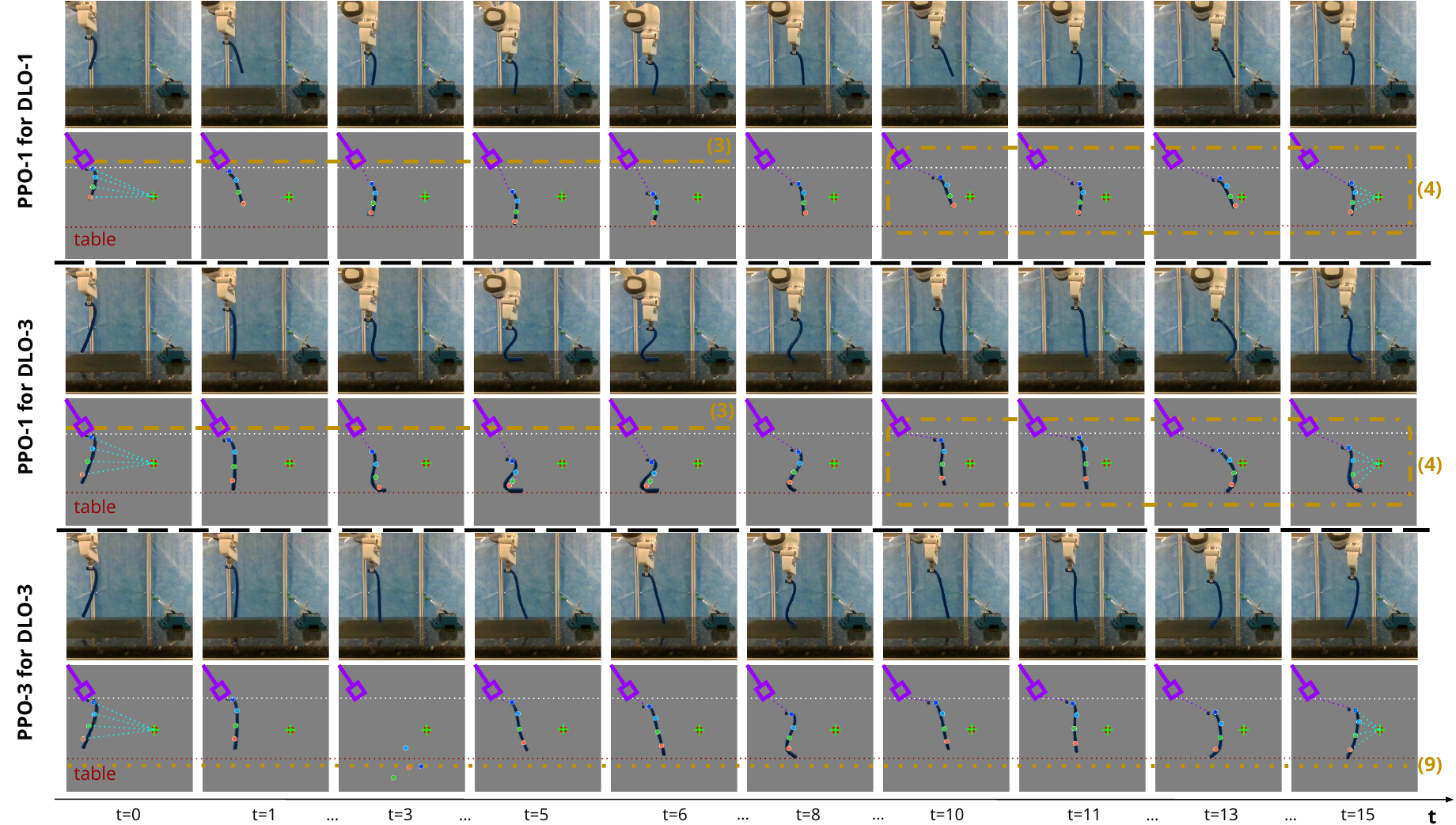}
    \caption{Timelapses of real-world PPO 1 and 3 policy evaluations using DLOs 1 and 3. In conjuction with Figure~\ref{fig:eef-deltas-per-eval-steps}, we extrinsically confirm reported observations \textcolor{olive}{\textbf{(3)}} (same $1/3$ of trajectory), \textcolor{olive}{\textbf{(4)}} (similar reaching pattern) and \textcolor{olive}{\textbf{(9)}} (moving higher above the table). Dotted white and red lines visually mark the initial EEF $z$ and table surface respectively. A purple indicator on top left of each image marks the initial EEF position, with dotted purple lines marking its visual displacement. Dotted cyan lines in $t=\{0, 15\}$ mark DLO body distance to target.} 
    \label{fig:extrinsic-eval-timelapse}
\end{figure*}

For each experiment, we report the mean trajectories resulting from the accumulation of commanded EEF position deltas (commanded \emph{actions}, $\langle dx, dz \rangle$) over $4$ repetitions.
These are presented in Fig.~\ref{fig:eef-deltas-per-eval-steps}, with shaded regions showing the standard deviation between different repetitions. We observe many interesting \textbf{motion patterns, which indicate adaptation of agent performance to the physical parameterisation of the respective evaluation DLO}. An indicative subset of these observations is extrinsically corroborated by Fig.~\ref{fig:extrinsic-eval-timelapse}. The following observations are highlighted in \textcolor{olive}{\textbf{olive}} annotations.
Policy underlines match plot line colours.

We see $\text{MoG}_{\text{DLO-0}}$ policy (\setulcolor{red}\ul{$\text{PPO-}0$}) showing the tightest ``roaming pattern'' \textcolor{olive}{\textbf{(1)}} to maximise the shorter and stiff DLO-0 reward close to the target. 
$\text{PPO-}0$ also follows the same first half of the trajectory \textcolor{olive}{\textbf{(2)}} for the very soft DLOs 1 and 3. 
Similarly, the $\text{MoG}_{\text{DLO-1}}$ policy (\setulcolor{green}\ul{$\text{PPO-}1$}), follows a very similar first third of the trajectory \textcolor{olive}{\textbf{(3)}} for the softer DLOs 1, 2 and 3, with some material-related variance, and then attempts a similar ``loop pattern'' for the last $6$ reaching steps with the softest DLOs 1 and 3 \textcolor{olive}{\textbf{(4)}}. 
In a related adaptation, we see that although both $\text{PPO-}1$ and $\text{MoG}_{\text{DLO-3}}$ policy (\setulcolor{cyan}\ul{$\text{PPO-}3$}) follow a similar pattern for the latter half of the trajectory for the shorter and stiff DLO-0, the PPO-1 pattern looks cleaner, due to being inherently conditioned on shorter DLOs \textcolor{olive}{\textbf{(5)}}.

For the $\text{MoG}_{\text{DLO-2}}$ policy (\setulcolor{blue}\ul{$\text{PPO-}2$}), we see a similar first half of the trajectory for the softest DLOs 1 and 3 (00-20), as well as for DLO-2 (00-50), although with greater variance \textcolor{olive}{\textbf{(6)}}. 
We also see a similar reaching pattern in the last $6$ steps for the longer DLOs 2 ($\text{270mm}$) and 3 ($\text{290mm}$) \textcolor{olive}{\textbf{(7)}}, with differences attributed to their different softness. 
PPO-2 and PPO-3 also exhibit a similar reaching pattern in the last $6$ steps for the shorter and softer DLO-1 \textcolor{olive}{\textbf{(8)}}. 
The most distinct sign of real-world adaptation to domain distribution occurs in PPO-3 trajectories maintaining a greater distance between the EEF and the table than any other DR policy \textcolor{olive}{\textbf{(9)}}.
For DLO-$\mu$, reasonably, $\text{PPO-}\mu$ is the most consistent (least variance). Notably, all simulated rollouts require a smaller sum of commanded $\langle dx, dz \rangle$ to approach the target, due to the \emph{perfect} dynamics of the robot and controller simulation.

\begin{figure}[t]
    \centering
    \includegraphics[width=1.0\columnwidth]{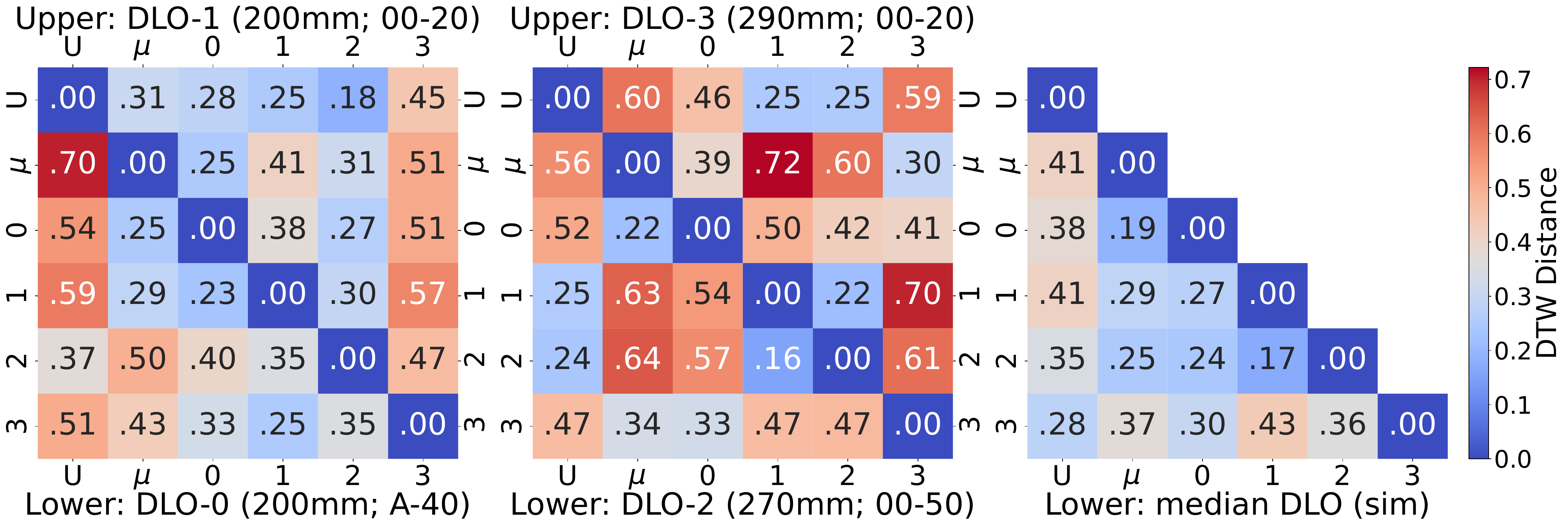}
    \caption{Dynamic Time Warping (DTW) heatmap of real and sim EEF trajectories (accum. commanded \emph{actions}) over PPO-$\{\mathit{U}, \mu, 0, 1, 2, 3\}$ evaluations.}
    \label{fig:eef-trajs-dtw-heatmap}
\end{figure}

Fig.~\ref{fig:eef-trajs-dtw-heatmap} shows the space of the mean EEF trajectories ($4$ rep. average) as a similarity heatmap, to illustrate the variation in the policy space for our $6$ policies evaluated over our $4$ real and $1$ simulated (median) DLOs. 
$\mathit{U}, \mu, 0, 1, 2, 3$ denote the respective PPO trajectories. 
For DLO-0, we see that the PPO-$\{0, 1\}$ trajectories, both conditioned on equally short $\text{200mm}$ DLOs, show the greatest similarity. 
For DLO-1, we see that the $\text{PPO-}\{\mathit{U}, 2\}$ trajectories are the most similar on average. 
For the longer DLOs 2 and 3, we have PPO-$\{1, 2\}$ being the most similar on average.
For DLO-$\mu$ (sim), PPO-$\{1, 3\}$ trajectories are the most different, followed by $\text{PPO-}\{\mathit{U}, \mu\}$ and $\text{PPO-}\{\mathit{U}, 1\}$. PPO-$\{1, 2\}$ are again the most similar. 
The greater similarity among all trajectories is due to the median parameters of the DLO, which shows the importance of DLO and domain distribution alignment.

Despite these extrinsic adaptation indications, Fig.~\ref{fig:eef-deltas-per-eval-steps} averaged rewards and pixel-distance to the target show relatively similar quantitative results, both in real and sim trajectories. This suggests that adaptation occurs at a behavioural level not fully captured by our sparse reward function. This highlights the \textbf{importance of trajectory-based evaluation in dynamic DLO manipulation}. Due to drag, inertia, and non-trivial DLO dynamics, success may require nuanced motion patterns that are difficult to capture by scalar proximity metrics.

\section{Discussion}
\label{sec:discussion}

In this paper, we extend the Real2Sim2Real literature with a proof of concept that an integrated distributional treatment of parameter inference, policy training, and zero-shot deployment can lead to significant agent behaviour adaptation over a set of parameterised DLOs. We use a prominent LFI method (BayesSim) to capture fine parameterisation differences of similarly shaped DLOs. Using the inferred MoG posteriors for DR during RL policy training may not lead to direct benefits in training performance (although this is largely task dependent); however, we observe strong object-centric agent performance indications during real-world deployment.

Through our framework practitioners of soft robotics and soft body manipulation can condition RL policies on system parameterisations inferred through vision, thus inducing object-centric performance, all the while operating in a shared action and observation space for both parameter inference and policy learning. In this context, a constraint to be addressed is that, although we reduce the reality gap in terms of \emph{realism} (observations), we do not yet account for \emph{physical accuracy} (true states). Thus, e.g., the inferred Young's of a DLO may not match the real value, especially if the Real2Sim algorithm has to account for higher-order $\boldsymbol{\theta}$.


\bibliographystyle{IEEEtran}
\bibliography{bibliography}

\end{document}